\documentclass{article}

\usepackage{arxiv}

\usepackage[utf8]{inputenc} 
\usepackage[T1]{fontenc}    
\usepackage{hyperref}       
\usepackage{url}            
\usepackage{booktabs}       
\usepackage{amsfonts}       
\usepackage{nicefrac}       
\usepackage{microtype}      
\usepackage{lipsum}
\usepackage{graphicx}
\usepackage{amssymb}
\graphicspath{ {./images/} }

\title{DECWA : Density-Based Clustering using Wasserstein Distance}

\author{
 Nabil El Malki \\
  IRIT\\
  \texttt{elmalki.nabil@gmail.com} \\
   \And
 Robin Cugny \\
  IRIT\\
  \texttt{robin.cugny@irit.fr} \\
  \And
 Olivier Teste \\
  IRIT\\
  \texttt{olivier.teste@irit.fr} \\
  \And
 Franck Ravat \\
  IRIT\\
  \texttt{franck.ravat@irit.fr} \\
}

\hypersetup{
pdftitle={DECWA : Density-Based Clustering using Wasserstein Distance},
pdfsubject={cs.LG},
pdfauthor={Nabil El Malki, Robin Cugny, Olivier Teste, Franck Ravat},
pdfkeywords={Density-based clustering, Wasserstein distance, Information system},
}

\begin{document}
\maketitle
\begin{abstract}
Clustering is a data analysis method for extracting knowledge by discovering groups of data called clusters. Among these methods, state-of-the-art density-based clustering methods have proven to be effective for arbitrary-shaped clusters. Despite their encouraging results, they suffer to find low-density clusters, near clusters with similar densities, and high-dimensional data. Our proposals are a new characterization of clusters and a new clustering algorithm based on spatial density and probabilistic approach. First of all, sub-clusters are built using spatial density represented as probability density function ($p.d.f$) of pairwise distances between points. A method is then proposed to agglomerate similar sub-clusters by using both their density ($p.d.f$) and their spatial distance. The key idea we propose is to use the Wasserstein metric, a powerful tool to measure the distance between $p.d.f$ of sub-clusters. We show that our approach outperforms other state-of-the-art density-based clustering methods on a wide variety of datasets.
\end{abstract}


\section{Introduction}
Clustering methods are popular techniques widely used to extract knowledge from a variety of datasets in numerous applications \cite{jain_survey_10}. Clustering methods aim at grouping similar data into a subset known as cluster. 
Formally, the clustering consists in partitioning a dataset annotated $X = \{x_1,...,x_n\}$ with $n=|X|$ into $c$ clusters $C_1, . . . , C_c$, so that $X = \cup_{i=1}^{c} C_i$ \cite{jain_survey_10}. 
 
In this paper, we only consider hard clustering \cite{THEODORIDIS2009701} according to which $\forall i \in [1..c], \forall j \ne i \in [1..c], C_i \cap C_j = \varnothing$.
We focus on density-based clustering methods \cite{densitybased} that are able to identify clusters of arbitrary shape. Another interesting element in these approaches is that they do not require the user to specify the number of clusters $c$. Density-based clustering is based on the exploration of high concentrations (density) of points in the dataset \cite{Aggarwal2013}. In density-based clustering, a cluster in a data space is a contiguous region of high point density separated from other such clusters by contiguous regions of low point density \cite{densitybased}. Density-based clustering has difficulties to detect clusters having low densities regarding high-density clusters.
Low-density points are either considered as outliers or included in another cluster. In the same way, near clusters of similar densities are often grouped in one cluster. Moreover, density-based clustering does not manage well in high-dimensional data \cite{Aggarwal2013} because the density is not evenly distributed and may vary severely. This paper tackles these challenging issues by defining a new density-based clustering approach. 

Among the most known density-based algorithms are DBSCAN \cite{dbscan}, OPTICS \cite{optics}, HDBSCAN \cite{hdbscan}, DBCLASD \cite{dbclasd}, and DENCLUE \cite{denclue2}. 
Historically first, DBSCAN introduces density as a minimum number of points within a given radius to discover clusters. 
Nevertheless, DBSCAN poorly manages clusters having different densities. OPTICS addresses these varying density clusters by ordering points according to a density measure. HDBSCAN improved the approach by introducing a new density measure and an optimization function aiming at finding the best clustering solution. Although these approaches solve part of the problem of varying density clusters, they suffer from unevenly distributed density and high-dimensional datasets. They still mismanage low-density clusters by tending to consider them as outliers or to merge them into a higher-density cluster.

DBCLASD introduces a probabilistic approach of the density. DBCLASD assumes that clusters follow a uniform probability law allowing it to be parameter-free \cite{denclue2}. However, it suffers from detecting non-uniform clusters because of this strong assumption. As DBCLASD is also a grid-based approach, its density calculations are less precise when the dimension increases \cite{Aggarwal2013}.

Finally, DENCLUE detects clusters using the probability density function ($p.d.f$) of points in data space. DENCLUE extends approaches of clustering such as DBSCAN \cite{dbscan} and K-means \cite{denclue2}. Therefore it also inherits from these the difficulty to detect low-density clusters when the density is not evenly distributed.

The common problems of the existing density-based clustering approaches are related to the difficulty of handling low-density clusters, near clusters of similar densities, and high-dimensional data. The other limit concerns their inefficiency to properly handle nested clusters of different shapes and uneven distribution of densities.
In this paper, we propose a new clustering approach that overcomes these limitations. Briefly, the contributions of this article are as follows:
(1) We propose DECWA (DEnsity-based Clustering using WAsserstein distance), a hybrid solution combining density and probabilistic approaches. It first produces sub-clusters using the $p.d.f$ defined on pairwise distances. Then, it merges sub-clusters with similar $p.d.f$ and close in distance.
(2) We propose to consider every cluster as a contiguous region of points where the $p.d.f$ has its own law of probability to overcome the previously explained limitations. 
(3) We conducted experiments on a wide variety of datasets, DECWA outperforms state-of-the-art density-based algorithms in clustering quality by an average of 20\%. Also, it works efficiently in high-dimensional data comparing to the others.

\section{Contribution}

To address the limits described earlier, we propose to consider a cluster as a contiguous region of points where density follows its own law of probability. Formally, we represent a cluster $C_i$ by a set of pairwise distances between the points contained in the cluster $C_i$. This set is annotated $D_i$ and follows any law of probability. $D_i$ is computed via any distance $d:X \times X \to \mathbb{R}^+$ that has to be at least symmetric and reflexive. 

The proposed solution consists of four consecutive steps: 1) The first step transforms the dataset $X$ into a $k$-nearest neighbor graph representation where nodes are points and edges are pairwise distances, $k$ is a hyperparameter. The graph is then reduced to its Minimum Spanning Tree (MST) in order to keep significant distances.
2) The second step consists in calculating the $p.d.f$ from the significant distances of the MST. This $p.d.f$ is used to determines the extrema, from which extraction thresholds are determined.
3) The third step consists of extracting sub-graphs from the MST according to extraction thresholds and identifying corresponding sub-clusters. 
4) The fourth step is to agglomerate sub-clusters according to spatial and probabilistic distances. We opt for the Wasserstein distance to measure the similarity between probability distributions. 

\subsection{Graph-oriented modeling of the dataset}\label{Graph_approach_section }

To estimate the distance $p.d.f$, we must first calculate distances between the points of the dataset. The computation strategy of pairwise distances is based on the $k$-nearest neighbor method \cite{Aggarwal2013}.

Therefore, the first step is the construction of the $k$-nearest neighbor graph of the dataset. From the dataset $X$, an undirected weighted graph annotated $G=(V,E)$ is constructed. $V$ is the set of vertices representing data points, and $E$ is the set of edges between data points. For each point, we determine $k$ edges to $k$ nearest neighbor points. The weight of each edge is the distance between the two linked points. 

To gain efficiency and accuracy, it is possible to get rid of many unnecessary distances. We favor short distances, avoiding keeping too large distances in the $k$-nearest neighbors. This idea tends to obtain dense (connected) sub-graphs.
To do so, we generate a minimum spanning tree (MST) from $G$.
An MST, denoted $G^{min}$, is a connected graph where the sum of the edge weights is minimal. Several algorithms exist to generate an MST, we used Kruskal \cite{kruskal_shortest_1956}.
As $k$ increases, the number of connections of each node in graph $G$ gets bigger too. Consequently, it also increases the number of possible solutions for Kruskal. As a result, the MST is more optimal. This might lead DECWA to better clustering results.

\subsection{Probability density estimation}\label{Density_ estimation_section}
The overall objective is to identify homogeneous regions where linked points have similar distances. We then estimate the distance $p.d.f$ to detect groups of points having a similar spatial density.

For estimating the $p.d.f$, we use the kernel density estimation ($KDE$) \cite{KDE}. Its interest lies in the fact that it makes no a priori hypothesis on the probability law of distances. The density function is defined as the sum of the kernel functions $K$ on every distance. The commonly used kernels are Gaussian, uniform and triangular.
The $KDE$ equation is below :

$${\widehat{f}}_{h}(a)={\frac{1}{(n-1)h}}\sum_{i=1}^{n-1}K{\Big (}{\frac{a-a_{i}}{h}}{\Big)}$$

In our case, $a$ and $a_i$ correspond to the distances contained in $G^{min}$. We have $n=|X|$ (number of nodes in $G^{min}$), and $n-1$ is the number of distances (number of edges in $G^{min}$).

The smoothing factor $h \in ]0 ; +\infty[$ is an hyperparameter, called bandwidth. It acts as a precision parameter, in our case, it influences the number of extrema detected in the $p.d.f$ and therefore, the number of different densities.  

\subsection{Graph division}

The overall objective is to separate different densities, hence, the next step is to find where to cut the $p.d.f$. The significantly different densities are detected with the maxima of the distance $p.d.f$ curve. In order to separate highly represented distances to less represented distances, we consider an extraction threshold as the mid-distance between each maximum and its consecutive minimum, and conversely between each minimum and its consecutive maximum. Mid-distances allow capturing regions that are difficult to detect (low-density regions, or dense regions containing few points). Another interesting consequence concerns overlapping regions that have the same densities. These regions are often very difficult to capture properly, whilst our mid-distance approach makes it possible to detect these cases because overlapping between regions induces a density variation.
Once the mid-distances are identified on the $p.d.f.$ curve, we apply a process to treat successively the list of mid-distances in descending order. We generate sub-clusters, from the nodes of $G^{min}$, and sets of distances of sub-clusters, from the edges of $G^{min}$.

From $G^{min}$, we extract connected sub-graphs where all the nodes that are exclusively linked by edges greater than the current mid-distance. A node linked to an edge greater than the mid-distance and another edge less than the mid-distance is not included.
A sub-cluster ($C_i$) is composed of points that belong to one connected sub-graph. For each sub-cluster, we produce its associated set of distances ($D_i$) from edges between nodes of the sub-graph.
A residual graph is made up of edges whose distances are less than the mid-distance, and all the nodes linked by these edges. This residual graph is used in the successive iterations. 
At the last iteration (i.e. the last mid-distance), the residual graph is also used to generate sub-clusters and associated sets of distances.
After this step, some sub-graphs are close to each other while having a similar $p.d.f$. According to our cluster characteristics, they should be merged. 

\subsection{Agglomeration of sub-clusters}

This step aims at generating clusters from sub-cluster agglomeration. The main difficulty of this step is to determine which sub-clusters should be merged. When the distances between two sub-clusters are close, it is difficult to decide whether or not to merge them. To arbitrate this decision, we use their density, represented as the distance $p.d.f$. Only sub-clusters with similar distance $p.d.f$ will be merged.
The agglomeration process consists in merging every sub-clusters $C_i$ and $C_j$ that satisfy two conditions. 

The first condition is that $C_i$ and $C_j$ must be spatially close enough. To ensure this, $d(C_i,C_j) \le \lambda$ must be respected, with $\lambda \in \mathbb{R^{+}}$ a hyperparameter.
It is necessary to determine the distance between sub-clusters ($d(C_i,C_j)$) to verify this condition. However, calculating every pairwise distance between two sub-clusters is a time-consuming operation. Therefore, we propose another solution based on $G^{min}$. We consider that the value of the edge linking sub-graphs corresponding to $C_i$ and $C_j$ as the distance between $C_i$ and $C_j$. Because of the MST structure, we assume that this distance is nearly the minimum distance between the points of $C_i$ and $C_j$.

The second condition relates to the similarity of the distance distributions $D_i$ and $D_j$. The purpose of this condition is to ensure $ws(D_i,D_j) \le \alpha$, with $\alpha \in \mathbb{R^{+}}$ a hyperparameter, and $ws$ the Wasserstein distance. We have opted for the Wasserstein \cite{optimaltransport} distance as a measure of similarity between probability distributions because it can capture small differences on similar density clusters. The values of $\alpha$ and $\lambda$ allow new  fusions as they increase, that tends to generate fewer clusters and conversely.

We introduce now an iterative process for merging sub-clusters. It consists in traversing $G^{min}$ by iterating on its edges. For each edge whose nodes are not in the same sub-clusters (e.g. $C_i$ and $C_j$), we verify that $d(C_i,C_j) \le \lambda$ and $ws(D_i,D_j) \le \alpha$. In this case, $C_i$ and $C_j$ are merged. This is operated by the union of $D_i$, and $D_j$ and considering the points of $C_i$ and $C_j$ as belonging to the same sub-cluster. The graph traversal of $G^{min}$ is repeated until there are no longer merges. The last iteration does not result in any merging.

\section{Experiments}

\begin{table*}[]
\caption{Experimental results}
\label{results}
\footnotesize
\begin{tabular}{p{1.85cm}p{1.2cm}p{0.5cm}p{0.3cm}p{0.3cm}p{0.3cm}p{1.1cm}p{0.3cm}p{1.1cm}p{0.3cm}p{1.1cm}p{0.3cm}l}
\toprule
                 & \multicolumn{2}{l}{} &             &                 & \multicolumn{2}{c}{\textbf{DECWA}} & \multicolumn{2}{c}{\textbf{DBCLASD}} & \multicolumn{2}{c}{\textbf{DENCLUE}} & \multicolumn{2}{c}{\textbf{HDBSCAN}} \\
\midrule
\textbf{Dataset} & dimensions  & size   & LD & SD & \textbf{ARI}    & outliers(\%)    & \textbf{ARI}      & outliers(\%)     & \textbf{ARI}      & outliers(\%)     & \textbf{ARI}      & outliers(\%)     \\
\midrule
twodiamonds      & 2           & 800    &             &                 & 0.99            & 0.01            & 0.95              & 0.02             & \textbf{1.00}     & \textbf{0.00}    & 0.98              & 0.01             \\
jain             & 2           & 373    &             &                 & \textbf{1.00}   & \textbf{0.00}   & 0.90              & 0.03             & 0.45              & 0.06             & 0.94              & 0.01             \\
cluto-t7.10k     & 2           & 10000  &             &                 & 0.93            & 0.03            & 0.79              & 0.06             & 0.34              & \textbf{0.00}    & \textbf{0.95}     & 0.10             \\
compound         & 2           & 399    & x           & x               & \textbf{0.93}   & \textbf{0.00}   & 0.77              & 0.06             & 0.82              & 0.05             & 0.83              & 0.04             \\
pathbased        & 2           & 300    &             & x               & \textbf{0.76}   & \textbf{0.00}   & 0.47              & 0.03             & 0.56              & 0.00             & 0.42              & 0.02             \\
iris             & 4           & 150    &             & x               & \textbf{0.84}   & 0.03            & 0.62              & 0.07             & 0.74              & \textbf{0.00}    & 0.57              & \textbf{0.00}    \\
cardiotocography & 35          & 2126   &             &                 & \textbf{0.52}   & \textbf{0.03}   & 0.24              & 0.17             & 0.47              & 0.02             & 0.08              & 0.28             \\
plant            & 65          & 1600   & x           & x               & \textbf{0.22}   & \textbf{0.00}   & 0.04              & 0.07             & 0.14              & 0.17             & 0.04              & 0.38             \\
GCM              & 16064       & 190    & x           & x               & \textbf{0.48}   & \textbf{0.03}   & 0.18              & 0.18             & 0.24              & 0.06             & 0.27              & 0.27             \\
new3             & 26833       & 1519   &             & x               & \textbf{0.41}   & \textbf{0.03}   & 0.09              & 0.45             & 0.08              & 0.13             & 0.13              & 0.27             \\
kidney\_uterus   & 10936       & 384    &             & x               & \textbf{0.81}   & \textbf{0.00}   & 0.53              & 0.08             & 0.56              & 0.09             & 0.53              & 0.26             \\
\midrule
\textbf{Average}          &             &        &             &                 & \textbf{0.72}   & \textbf{0.02}   & \textbf{0.51}     & \textbf{0.11}    & \textbf{0.49}     & \textbf{0.05}    & \textbf{0.52}     & \textbf{0.15} \\
\bottomrule
\end{tabular}
\end{table*}

We conducted an experimental study to show the effectiveness and the robustness of DECWA compared to state-of-the-art density-based methods. It was applied to a variety of synthetic and real datasets, using different distances depending on the data. The synthetic datasets, as well as the Decwa advantages presentation, are available on this link\footnote{https://github.com/nabilEM/DECWA}.

\subsection{Experiment protocol}

\subsubsection{Algorithms.} DECWA was compared to DBCLASD \cite{dbclasd}, DENCLUE \cite{denclue2} and HDBSCAN \footnote{The hierarchical structure proposed by HDBSCAN is exploited by the Excess Of Mass (EOM) method to extract clusters, as used by the authors in \cite{hdbscan}.} \cite{hdbscan}. For the experiments conducted in this paper, DECWA used Gaussian kernel in division phase and performed only one graph traversal in the agglomeration phase.
Algorithm parameter values are defined through the repetitive
application of the random search approach (1000 iterations). This, in order to obtain the best score that the four algorithms can have. DBCLASD (parameter-free and incremental) was subject to the same approach but by randomizing the order of the data points because it is sensitive to it. 
Clustering quality is measured by the commonly used metric named Adjusted Rand Index ($ARI$) \cite{hubert1985comparing}. In addition, we also report the ratio of outliers produced by each algorithm for each dataset. 

\subsubsection{Synthetic datasets.} 
The two-dimensional diamond, jain, cluto-t7.10k, compound, and pathbased datasets are synthetic. They contain clusters of different shapes and densities. For these data, the Euclidean distance was used. 
\subsubsection{Real datasets.} The cardiotocography dataset \cite{cardio} is a set of 2126 fetal cardiotocogram records, with 35 attributes providing vital information on the state of the fetus, grouped into 10 classes. Canberra distance was applied to it. 
Plant dataset \cite{plant} consists of 1600 leaves of 100 different classes. Each leaf is characterized by 64 shape measurements retrieved from its binary image. The Euclidean distance was used on Plant. 
Another known dataset, Iris \cite{Dua:2019} (150 iris flowers, 4 dimensions, 3 classes) was used with the Manhattan distance. 
A collection of two very large biological datasets were tested. The first dataset, GCM \cite{Ramaswamy15149}, is used to diagnose the type of cancer (14 classes). It consists of 190 tumor samples. Each one represented by the expression levels of 16063 genes. The second dataset (kidney$\_$uterus) \cite{gene} consists of 384 tumor samples to be classified into two classes (kidney or uterus). Each sample is the expression level of 10937 genes. Given that the attributes all correspond to the same nature (gene expression) then the Bray-Curtis distance was applied to it. 
New3 \cite{10.1007/3-540-45372-5_46}, a very high-dimensional dataset, is a set of 1519 documents (6 classes). Each document is represented by the term-frequency vector of size 26833.The cosine distance was used to measure the similarity between documents. 

\subsection{Results and discussion}
The results are reported in the table \ref{results} (best scores are in bold). $LD$ column stands for low-density, it means that the datasets have at least one low-density cluster comparing to others. $SD$ stands for near clusters of similar densities, it means that the marked datasets have at least two overlapping clusters with similar densities. These were detected according to the intra-cluster and inter-cluster distance using ground truth.

In many cases, DECWA outperforms competing algorithms by a large margin on average 20\% (e.g. \textit{jain} dataset, ARI margin is $1-0.45=0.55$). DECWA has the best results in datasets with a low-density cluster (e.g. \textit{compound} and \textit{GCM}). In this case, there is an ARI margin of 21\% on average in favor of DECWA. Though datasets with very high dimensions are problematic for the other algorithms, DECWA is able to give good results. Indeed, there is an ARI margin of 30\% on average in favor of DECWA for the last three datasets.
Near clusters of similar densities are also correctly detected by DECWA. Some datasets like \textit{iris, kidney$\_$uterus} and \textit{pathbased} have overlapping clusters and yet DECWA separates them correctly. There is an ARI margin of 25\% in favor of DECWA for datasets having this problematic. 

The outlier ratio is not relevant in case of a bad ARI score because in this case, although the points are placed in clusters, clustering is meaningless. DECWA is the one that returns the least outliers on average while having a better ARI score.

We conducted a statistical study, as recommended in \cite{demsar}, to confirm the significant difference in performance between the algorithms and the robustness of DECWA comparing to the others. The overall concept of the study has two steps. 1) First, a statistical test (Iman-Davenport\cite{iman}) is performed to determine if there is a significant difference between the algorithms at the ARI level.
2) If so, a pairwise comparison of algorithms is performed, via a post-hoc test (Shaffer \cite{shaffer}), to identify the gain of one method over the others. Both tests return a p-value (in the case of the second test, it is returned for each comparison). The p-value allows us to decide on the rejection of the hypothesis of equivalence between algorithms. To reject the hypothesis, the p-value must be lower than a significance threshold $s$ that we set at 0.01.  
The p-value by returned the first test is $5.38e^{-5}$. It is significantly small compared to $s$. This means that the algorithms are different in terms of ARI performance. 
Second, these differences are analyzed by the Shaffer Post-hoc test. It returns a p-value for each test on a pair of algorithms. Indeed, for the case of DECWA, the p-value is much lower than $s$ when comparing DECWA to others ($3.6e^{-4}$ with DBCLASD, $3.0e^{-3}$ with DENCLUE and $1.0e^{-3}$ with HDBSCAN), which proves that DECWA is significantly different from the others. For the others, the p-value is equal to $1.0$ in all the tests concerning them, which statistically shows that they are equivalent. The ranking of the algorithms according to the ARI was done by Friedman's aligned rank \cite{garcia}. DECWA is ranked as the best. All in all, DECWA is significantly different from the others and is the best performing.

\subsection{CONCLUSION AND PERSPECTIVES}
We proposed DECWA, a clustering algorithm based on spatial and probabilistic density adapted to high-dimensional data. We introduced a new cluster characterization to allow efficient detection of clusters with different densities. Experiments were performed on various datasets and we showed statistically that DECWA outperforms state-of-the-art density-based methods. Our future research integrates the application of DECWA in specific domains and on complex data as multidimensional time series.

\bibliographystyle{unsrt}  

\bibliography{references}

\begin{thebibliography}{10}

\bibitem{jain_survey_10}
Anil Jain.
\newblock Data clustering: 50 years beyond k-means.
\newblock {\em Pattern Recognition Letters}, 31:651--666, 06 2010.

\bibitem{THEODORIDIS2009701}
Sergios Theodoridis and Konstantinos Koutroumbas.
\newblock Chapter 14 - clustering algorithms iii: Schemes based on function optimization.
\newblock In {\em Pattern Recognition (Fourth Edition)}, pages 701 -- 763. Academic Press, fourth edition edition, 2009.

\bibitem{densitybased}
Hans-Peter Kriegel, Peer Kröger, Joerg Sander, and Arthur Zimek.
\newblock Density-based clustering.
\newblock {\em Wiley Interdisc. Rew.: Data Mining and Knowledge Discovery}, 1:231--240, 05 2011.

\bibitem{Aggarwal2013}
Charu~C. Aggarwal and Chandan~K. Reddy.
\newblock {\em Data Clustering: Algorithms and Applications}.
\newblock Chapman \& Hall/CRC, 1st edition, 2013.

\bibitem{dbscan}
Martin Ester, Hans-Peter Kriegel, Joerg Sander, and Xiaowei Xu.
\newblock A density-based algorithm for discovering clusters in large spatial databases with noise.
\newblock volume~96, pages 226--231, 01 1996.

\bibitem{optics}
Mihael Ankerst, Markus Breunig, Hans-Peter Kriegel, and Joerg Sander.
\newblock Optics: Ordering points to identify the clustering structure.
\newblock volume~28, pages 49--60, 06 1999.

\bibitem{hdbscan}
Ricardo Campello, Davoud Moulavi, and Joerg Sander.
\newblock Density-based clustering based on hierarchical density estimates.
\newblock volume 7819, pages 160--172, 04 2013.

\bibitem{dbclasd}
Xiaowei Xu, Martin Ester, Hans-Peter Kriegel, and Joerg Sander.
\newblock A distribution-based clustering algorithm for mining in large spatial databases.
\newblock pages 324--331, 01 1998.

\bibitem{denclue2}
Alexander Hinneburg and Hans-Henning Gabriel.
\newblock Denclue 2.0: Fast clustering based on kernel density estimation.
\newblock volume 4723, pages 70--80, 09 2007.

\bibitem{kruskal_shortest_1956}
Joseph~B. Kruskal.
\newblock {On the Shortest Spanning Subtree of a Graph and the Traveling Salesman Problem}.
\newblock {\em Proceedings of the American Mathematical Society}, 1956.

\bibitem{KDE}
Richard~A. Davis, Keh-Shin Lii, and Dimitris~N. Politis.
\newblock {Remarks on Some Nonparametric Estimates of a Density Function}.
\newblock In {\em Selected Works of Murray Rosenblatt}, pages 95--100. Springer New York, 2011.

\bibitem{optimaltransport}
C\'edric Villani.
\newblock {\em {Optimal transport : old and new}}.
\newblock Springer, 2009.

\bibitem{hubert1985comparing}
L.~Hubert and P.~Arabie.
\newblock {Comparing partitions}.
\newblock {\em Journal of classification}, 2(1):193--218, 1985.

\bibitem{cardio}
Diogo Ayres-de Campos, João Bernardes, Antonio Garrido, Joaquim Marques-de Sá, and Luis Pereira-Leite.
\newblock Sisporto 2.0: A program for automated analysis of cardiotocograms.
\newblock {\em The Journal of Maternal-Fetal Medicine}, 9(5):311--318, 2000.

\bibitem{plant}
Charles Mallah, James Cope, and James Orwell.
\newblock Plant leaf classification using probabilistic integration of shape, texture and margin features.
\newblock {\em Pattern Recognit. Appl.}, 3842, 02 2013.

\bibitem{Dua:2019}
Dheeru Dua and Casey Graff.
\newblock {UCI} machine learning repository, 2017.

\bibitem{Ramaswamy15149}
Sridhar Ramaswamy, Pablo Tamayo, Ryan Rifkin, Sayan Mukherjee, Chen-Hsiang Yeang, Michael Angelo, Christine Ladd, Michael Reich, Eva Latulippe, Jill~P. Mesirov, Tomaso Poggio, William Gerald, Massimo Loda, Eric~S. Lander, and Todd~R. Golub.
\newblock Multiclass cancer diagnosis using tumor gene expression signatures.
\newblock {\em PNAS}, 98:15149--15154, 2001.

\bibitem{gene}
Gregor Stiglic and Peter Kokol.
\newblock Stability of ranked gene lists in large microarray analysis studies.
\newblock {\em Journal of biomedicine \& biotechnology}, 2010:616358, 06 2010.

\bibitem{10.1007/3-540-45372-5_46}
Eui-Hong Han and George Karypis.
\newblock Centroid-based document classification: Analysis and experimental results.
\newblock In {\em Proceedings of the 4th European Conference on Principles of Data Mining and Knowledge Discovery}, page 424–431. Springer-Verlag, 2000.

\bibitem{demsar}
Janez Demsar.
\newblock Statistical comparisons of classifiers over multiple data sets.
\newblock {\em Journal of Machine Learning Research}, 7:1--30, 01 2006.

\bibitem{iman}
David~J. Sheskin.
\newblock {\em Handbook of Parametric and Nonparametric Statistical Procedures}.
\newblock Chapman \& Hall/CRC, 4 edition, 2007.

\bibitem{shaffer}
Juliet Shaffer.
\newblock Modified sequentially rejective multiple test procedures.
\newblock {\em Journal of The American Statistical Association}, 81:826--831, 09 1986.

\bibitem{garcia}
Salvador Garía and Francisco Herrera.
\newblock An extension on "statistical comparisons of classifiers over multiple data sets" for all pairwise comparisons.
\newblock {\em Journal of Machine Learning Research - JMLR}, 9, 12 2008.

\end{thebibliography}

\end{document}